\documentclass[letterpaper, 10 pt, journal, twoside]{IEEEtran}  

\IEEEoverridecommandlockouts                              

\usepackage{amsmath} 
\usepackage{amssymb}  
\usepackage{graphicx}
\usepackage{cite}
\usepackage{tikz}
\usepackage{tikz-qtree}
\usepackage{algorithm}
\usepackage[noend]{algpseudocode}

\usetikzlibrary{matrix,calc,positioning,decorations.pathreplacing}

\title{\LARGE\bf iBoW-LCD: An Appearance-based Loop Closure Detection \\Approach using Incremental Bags of Binary Words*}

\author{Emilio Garcia-Fidalgo and Alberto Ortiz
\thanks{This work was supported by the EU H2020 project ROBINS (GA 779776).}
\thanks{All authors are with the Department of Mathematics and Computer Science, University of the Balearic Islands, 07122 Palma, Spain. {\tt\small \{emilio.garcia, alberto.ortiz\}@uib.es}.}%
\thanks{* Copyright~\copyright{} IEEE 2018 All rights reserved. IEEE Robotics and Automation Letters (2018). Digital Object Identifier (DOI): 10.1109/LRA.2018.2849609.}
}

\begin{document}

\newcommand{\change}[1]{{#1}}

\maketitle
\thispagestyle{empty}
\pagestyle{empty}

\begin{abstract}
In this paper, we introduce iBoW-LCD, a novel appearance-based loop closure detection method. The presented approach makes use of an incremental Bag-of-Words (BoW) scheme based on binary descriptors to retrieve previously seen similar images, avoiding any vocabulary training stage usually required by classic BoW models. In addition, to detect loop closures, iBoW-LCD builds on the concept of \emph{dynamic islands}, a simple but effective mechanism to group similar images close in time, which reduces the computational times typically associated to Bayesian frameworks. Our approach is validated using several indoor and outdoor public datasets, taken under different environmental conditions, achieving a high accuracy and outperforming other state-of-the-art solutions.
\end{abstract}


\vspace{-3mm}

\section{Introduction}
\label{sec:intro}
One of the most important aspects of Simultaneous Localization and Mapping (SLAM)~\cite{Durrant-Whyte2006} is to correctly manage the perceived information from the environment. Irrespective of the kind of sensor involved, there always intervene unavoidable noise sources that produce inaccurate measurements, leading to inconsistent representations when only raw sensor data is considered. \change{For this} reason, SLAM algorithms usually rely on \emph{loop closure detection} mechanisms, which entail the correct identification of previously visited places. A robust loop closure detection scheme leads to additional constraints for the map generation process, resulting into more consistent representations. Although a variety of sensors have been used for loop closure detection, in the last decades, a high number of visual solutions have emerged, specially motivated by the low cost of cameras, the increase in computing power and the richness of the sensor data provided. Using a camera as the main source of information to undertake the association problem is generically known as \emph{appearance-based} loop closure detection~\cite{Cummins2008a,Cummins2011,Angeli2008b,Galvez-Lopez2012,Garcia-Fidalgo2014ETFA,Khan2015,Zhang2016,Bampis2017}.

The performance of an appearance-based loop closure detection algorithm is highly influenced by the method used to describe the input images and the ability to retrieve previous images similar to the current one. Regarding image description, recent binary descriptors, such as BRIEF~\cite{Calonder2010}, ORB~\cite{Rublee2011}, LDB~\cite{Yang2014} or AKAZE~\cite{Alcantarilla2011}, are progressively replacing the classical real-valued descriptors like SIFT~\cite{Lowe2004} or SURF~\cite{Bay2006a}, given their reduced storage needs and computational times. \change{As for the next issue, image indexing,} the Bag of Words (BoW) model~\cite{Sivic2003,Nister2006} has proven to be an effective solution, specially when used in combination with an inverted index. In this model, the set of detected local features \change{is} quantized according to a set of representative features called \emph{visual words}, which conform a \emph{visual vocabulary}, from which a histogram of visual word occurrences can be derived as the image descriptor. This visual vocabulary is typically generated off line. \change{As a main limitation, these approaches need a training phase, which, depending on the number of descriptors (required to be high for an adequate performance) and the clustering technique used, can take a long time. Furthermore, this visual vocabulary is intended to be useful for generic scenarios, perhaps with a different appearance with regard to the training set, what can lead to additional false detections~\cite{Nicosevici2012}}. \change{In these cases, the vocabulary can be regenerated using images taken from the current environment, at the expense of a priori knowledge and more computation time. An alternative to cope with these issues is to build the dictionary in an incremental manner.}

This paper proposes a novel and effective method for \change{visual} loop closure detection \change{oriented to view/place recognition} called iBoW-LCD (\emph{Incremental Bag-of-Words Loop Closure Detection}). Our approach adopts an incremental Bag of Binary Words strategy, which is able to build a visual vocabulary in an on-line manner, avoiding the drawbacks of off-line approaches. This scheme, used in combination with an inverted index, is employed to efficiently retrieve previously seen images. A robust loop closure detection method is proposed next. It extends and enhances the concept of \emph{island}~\cite{Galvez-Lopez2012} in order to avoid images taken from the same place to compete among them as loop closure candidates. iBoW-LCD is validated using different public indoor and outdoor datasets and compares favourably against several state-of-the-art solutions, outperforming them in several ways.

\change{Regarding the BoW strategy, our previous works~\cite{Garcia-Fidalgo2014ETFA,Garcia-Fidalgo2016,Garcia-Fidalgo2017,GarciaFidalgo2018Book} adopted a purely incremental approach, where the visual words were never forgotten (removed) but increased in number as new images were processed. In this work, we consider not only adding but also deleting words as they are not deemed useful by an optimized version of the incremental BoW approach. This results in similar performance (as shown by the experimental results) with significantly less visual words. Regarding loop closure, our previous works were mainly based on Bayes filtering, which usually exhibits increasing processing times as more images are considered by the filter. In this work, we replace that filter by a simpler but effective mechanism to visually close loops on the basis of the novel concept of \emph{dynamic island}, obtaining similar performance but reducing processing times. Unlike in~\cite{Garcia-Fidalgo2014ETFA}, the method presented in this paper is not based on the FLANN implementation of the Muja's algorithm~\cite{Muja2009} and it has been developed from scratch.}

The rest of the paper is organized as follows: Section~\ref{sec:relwork} \change{overviews} most important works in the field; Section~\ref{sec:ibow} introduces \change{the new incremental Bag of Binary Words scheme}; Section~\ref{sec:lcd} presents the loop closure detection approach; Section~\ref{sec:exresults} reports on the results obtained; and, finally, Section~\ref{sec:conclusions} concludes the paper and \change{discusses topics for future research}.

\section{Related Work}
\label{sec:relwork}
Most appearance-based loop closure detection solutions developed during the last years can be mainly classified according to the method used to describe the input images~\cite{Garcia-Fidalgo2015RAS}. In this respect, \change{some authors} have opted for using a holistic approach to compute a global descriptor of the image. This kind of descriptors are usually fast to compute, but less tolerant to illumination and view-point changes, what reduces their discriminative capabilities. In order to alleviate this effect, loop closure techniques based on global descriptors tend to match sequences instead of single images~\cite{Milford2012,Arroyo14IV,Arroyo14IROS}. This has been proven to be more robust against appearance changes, but renouncing to other desirable properties to detect loops, such as rotation invariance.


\change{In this line, CNN-based solutions~\cite{Sunderhauf2015a,Sunderhauf2015b,Arroyo2016,Arandjelovic2016} have recently emerged as effective against environmental changes. As a pioneering work, S\"{u}nderhauf et al.~\cite{Sunderhauf2015a} evaluated the utility of ConvNets for place recognition. In~\cite{Sunderhauf2015b}, they combined an object proposal technique with CNN features to match places over extreme appearance changes. Arroyo et al.~\cite{Arroyo2016} proposed a method where they fused the information from different convolutional layers to perform topological localization. In a recent work, Arandjelovic et al.~\cite{Arandjelovic2016} introduced a CNN architecture mainly based on a layer inspired in the VLAD image representation for weakly supervised place recognition. Despite the good performance shown by this kind of solutions, they are still disconnected from real SLAM and loop closure detection problems, as stated in~\cite{Bampis2017}.}

The BoW model~\cite{Sivic2003,Nister2006} is, by far, the most used technique for appearance-based loop closure, given its demonstrated efficiency for \change{retrieving} previous similar images. Solutions based on this scheme can be mainly classified as off-line and on-line, depending on the nature of the vocabulary building process. Key works that fall into the off-line category are the FAB-MAP algorithm~\cite{Cummins2008a} and its extension FAB-MAP 2.0~\cite{Cummins2011}, where a Chow-Liu tree was used to approximate the probabilities of visual word co-occurrences. G\'alvez-L\'opez and Tard\'os~\cite{Galvez-Lopez2012} trained a visual vocabulary based on BRIEF~\cite{Calonder2010}, promoting the use of binary descriptors for place recognition tasks. Using this vocabulary as a basis, they introduced a loop closure detection method based on the concept of \emph{islands} to group similar images close in time. \change{The authors prevent images with a similar appearance to compete among them as loop closure candidates splitting the image sequence into fixed-size intervals. The algorithm establishes a relationship between the query image and each island according to a global score, computed as the sum of the individual scores of each image belonging to the island.} \change{In this work, we extend this idea by adapting the generation of the islands to the operating environment,} \change{allowing islands of different and dynamic sizes,} as will be shown later. Mur-Artal and Tard\'os~\cite{Mur-Artal2014} enhanced their original algorithm~\cite{Galvez-Lopez2012} by using ORB~\cite{Rublee2011}, more robust against scale and rotation changes. A more recent work~\cite{Bampis2017} proposes an extension of the BoW model that groups visual words with similar optical flow when observed along two consecutive frames. These groups are called Structure-Aware and Viewpoint-Invariant High-Order Visual-Words (SVHVs). They naturally include the environment structure into the image description. \change{All the methods surveyed so far require a training stage. In this work, we want to address the problem from a different point of view, by building the dictionary in an on-line manner.}

Several on-line BoW attempts can be found in the literature~\cite{Angeli2008b,Labbe2013,Nicosevici2012,Garcia-Fidalgo2014ETFA,Khan2015,Zhang2016}. Among them, the work by Angeli et al.~\cite{Angeli2008b}, which proposes a loop closure method based on an incremental BoW scheme~\cite{Filliat2007} and a Bayesian filtering framework, can be considered of high importance in the field. \change{Other on-line approaches involve} RTAB-Map~\cite{Labbe2013} and OVV~\cite{Nicosevici2012}, \change{although these approaches are based on real-valued descriptors}. Recently, an incremental BoW scheme based on binary descriptors called IBuILD~\cite{Khan2015} was introduced. \change{This work describes a method to construct a visual dictionary in an on-line manner, aiming at loop closure detection}. However, as stated by the authors, their approach does not employ an indexing scheme for an efficient search of words, which affects the scalability of the algorithm. \change{The approach proposed in our paper features a hierarchical and incremental structure for such purpose, reducing the complexity during the BoW assignment process}. \change{In a more recent solution, Zhang et al.~\cite{Zhang2016} proposed a technique for learning a visual word from a pair of matched features along two consecutive frames. The learned descriptor has perspective invariance to camera motion. This technique is finally integrated into the IBuILD algorithm, which, as mentioned before, lacks of a hierarchical structure to efficiently search for visual words.}

\section{Incremental BoW for Image Indexing}
\label{sec:ibow}
In order to manage an increasing number of visual words, an efficient indexing scheme is required, since a linear search becomes infeasible. Normally, this problem is solved using hierarchical structures such as kd-trees~\cite{Silpa2008} or hierarchical k-means trees~\cite{Muja2009}, but these methods are not suitable for binary descriptors because they expect that the descriptor components can be continuously averaged. \change{Instead, hashing techniques~\cite{Gionis1999,Salakhutdinov2009} can be used for matching binary descriptors. In this respect, Muja and Lowe introduced in~\cite{Muja2012} a novel method that achieves better performance than hashing approaches. Furthermore, their method involves a hierarchical tree which is a perfect structure for adding and deleting descriptors, as it is required in our case.} In this work, we extend this method to be used as an incremental visual dictionary, as explained in the following sections.

\subsection{Overview of Muja's Approach}
\label{ssec:muja}
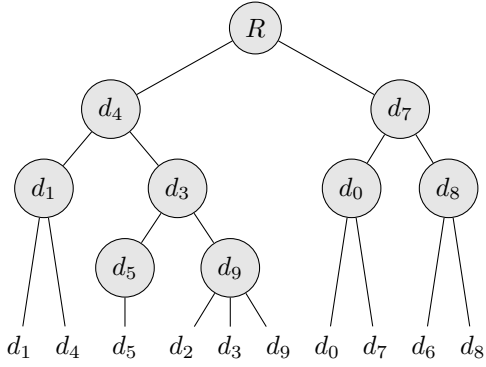
\begin{figure}[tb]
\centering
  \begin{tikzpicture}[scale=1.0, every node/.style={transform shape}]
  \tikzset{
    every internal node/.style=
      {
        circle,draw,fill=gray!20,align=center
      },
    frontier/.style=
      {
        distance from root=120pt
      },
    edge from parent path=
      {
        (\tikzparentnode) -- (\tikzchildnode)
      }
  }
  \Tree [
    .$R$
    [.$d_4$
      [.$d_1$
        $d_1$
        $d_4$
      ]
      [.$d_3$
        [.$d_5$
          $d_5$
        ]
        [.$d_9$
          $d_2$
          $d_3$
          $d_9$
        ]
      ]
    ]
    [.$d_7$
      [.$d_0$
        $d_0$
        $d_7$
      ]
      [.$d_8$
        $d_6$
        $d_8$
      ]
    ]
  ]
  \end{tikzpicture}
  \vspace{-3.0mm}
  \caption{A simple example of a hierarchical tree built by means of the Muja's approach~\cite{Muja2012} to index 10 visual descriptors ($d_0, \ldots, d_9$) using as parameters $K=2$ and $S=3$. Labels in non-leaf nodes (grey circles) indicate the descriptor selected randomly as the cluster centre during the building process.}
  \label{fig:tree}
\end{figure}
Muja and Lowe introduced an effective hierarchical structure~\cite{Muja2012} to index and match binary features, which requires less storage space and scales better than other hashing methods. This structure consists in a tree where non-leaf nodes contain cluster centres and leaf nodes store visual descriptors to be matched. \change{The visual words of the incremental vocabulary are hence stored in the leaf nodes.} To build one of these trees, the algorithm randomly selects, from the initial set of points, $K$ descriptors as cluster centres. Next, each remaining input descriptor is assigned to its closest cluster centre according to their Hamming distance. This process is repeated recursively until the number of descriptors within a cluster is below a certain threshold $S$. The authors also demonstrated that building several trees $T_i$ and using them in parallel during the search led to higher performance. An example of a tree built using this process is shown in Fig.~\ref{fig:tree}.

In order to search descriptors in parallel using several trees, the algorithm starts with a single traverse of each tree from the root until reaching a leaf node, selecting at each step the node closest to the query descriptor and adding the unexplored nodes to a priority queue. When a leaf node is reached, all the points within this node are linearly searched. After exploring each tree once, the search continues from the closest node stored in the priority queue. The process finishes when a certain amount of descriptors \change{has been examined (64 in our experiments)}.

\subsection{Visual Vocabulary Update}
\label{ssec:vvupd}
Muja's scheme was originally devised to index a static set of descriptors. Given that in our approach we handle an incremental visual dictionary, several modifications over the original approach have been introduced. First of all, binary descriptors are matched and combined during navigation to update the visual words of the vocabulary by means of a merging policy. As in our previous works~\cite{Garcia-Fidalgo2014ETFA,Garcia-Fidalgo2017}, we make use of a bitwise AND operation, which experimentally provides us better recognition performance than a bitwise OR. Formally stated:
\begin{equation}
B_{w_i}^{t} = B_{w_i}^{t-1} \land B_q\,,
\label{eq:merge}
\end{equation}
where $B_{w_i}^{t-1}$ is the binary descriptor associated to the visual word $w_i$ at time $t-1$, $B_q$ is a query descriptor and $B_{w_i}^{t}$ is the descriptor associated to the visual word $w_i$ after the fusion of descriptors. In~\cite{Garcia-Fidalgo2017}, we report on several experiments that demonstrate that this policy does not end up into degenerated descriptors (e.g. almost all bits set to zero).

Secondly, descriptors \change{without a match} in the index are incorporated into the dictionary as new visual words. \change{To this end, each descriptor is searched from the root until reaching a leaf node. Next, we assess if adding the corresponding new descriptor to the selected leaf node exceeds the maximum leaf size $S$. If that is the case, the node is recursively rebuilt adding the query descriptor to the original descriptor set. Otherwise, the descriptor is simply appended to the leaf node. Algorithm~\ref{alg:insert} illustrates this process.}
\begin{algorithm}[tb]
\caption{Adding a descriptor as a new visual word}
\label{alg:insert}
\begin{algorithmic}[1]
\Require{$T$: Hierarchical tree, $B$: Binary descriptor}
\State $node \leftarrow$ searchDescriptor($T, B$)
\If{numDescriptors($node$) + 1 $ < S$}
  \State appendDescriptorToNode($node$, $B$)
\Else
  \State $D \leftarrow$ getDescriptors($node$)
  \State $D = D \cup B$
  \State buildNodeRecursively($node, D$)
\EndIf
\end{algorithmic}
\end{algorithm}

Third, we maintain an inverted index. It stores, for each visual word, a list of images where it was found. Initially, the visual dictionary is created with the binary descriptors of the first image as a set of visual words. When an input image is processed, its extracted descriptors are matched against the visual words of the index applying the ratio test~\cite{Lowe2004}. \change{Matched descriptors are merged with their corresponding visual words using Eq.~\ref{eq:merge}}. Unmatched descriptors are added as temporary visual words to the vocabulary. In order to reduce the complexity of the index, these temporal visual words survive only if, after several consecutive frames $P_f$, they have been observed (e.g. matched) at least a certain number of times $P_o$. The inverted index is updated accordingly. The main purpose of this policy is to determine visual words that are most likely to be observed again if the agent returns to a previous location.

Lastly, we have provided the vocabulary with a mechanism to delete visual words to support the update policy outlined above. \change{After deleting a descriptor from the dictionary, the node where it was appended and its ancestors are recursively revised to assess if they still contain children nodes. A node without any children node is no longer required, and therefore, deleted. If the deleted descriptor coincides with the cluster centre, a new centre is randomly selected}. The process is summarized in Alg.~\ref{alg:delete}.
\begin{algorithm}[tb]
\caption{Deleting a visual word}
\label{alg:delete}
\begin{algorithmic}[1]
\Require{$T$: Hierarchical tree, $B$: Binary descriptor}
\State $node \leftarrow$ getNodeOfDescriptor($T, B$)
\State deleteDescriptor($node, B$)
\If{numDescriptors($node$) $ > 0$}
  \If{$B ==$ getClusterCentre($node$)}
    \State selectNewClusterCentre($node$)
  \EndIf
\Else
  \State $node_r \leftarrow$ getRootNode($node$)
  \State deleteChildNode($node_r, node$)
  \State deleteNodesRecursively($node_r$)
\EndIf
\end{algorithmic}
\end{algorithm}

\subsection{Retrieval of Similar Images}
\label{ssec:simimgs}
The approach introduced in this section is used to efficiently retrieve previous images which are similar to the current image. The inverted index allows us to efficiently compare a query image only with those images that share some visual words with it. A similarity score $s(I_t, I_k)$ is initialized to 0 for all possible $k$ previously seen frames. Being $z_t$ the set of binary descriptors extracted from the current image $I_t$, we search each descriptor of $z_t$ in the dictionary to find the closest visual word. Next, we obtain, from the inverted index, the list of images where this visual word has appeared, and add a weight to the score $s$ corresponding to each of the retrieved images. This weight is related to the term frequency - inverse document frequency (tf-idf~\cite{Sivic2003,Garcia-Fidalgo2017}) scoring, which reflects the importance of a visual word with regard to the visual vocabulary and the current image. After processing all descriptors in $z_t$, the ordered list of scores $s$ is returned as the images most similar to $I_t$. The source code of this image indexing method is available to the community\footnote{http://github.com/emiliofidalgo/obindex2} as \emph{OBIndex2} (Online Binary Index 2).

\section{Loop Closure Detection}
\label{sec:lcd}
This section details iBoW-LCD, a novel loop closure detection approach which makes use of the aforementioned OBIndex2. The source code is also available on line\footnote{http://github.com/emiliofidalgo/ibow-lcd}.

\subsection{Searching for Previous Images}
\label{ssec:llccands}
Given an image $I_t$ at time stamp $t$, the process starts querying the image index, as explained in Sec.~\ref{ssec:simimgs}. A buffer is used to store the most recent $p$ images, and hence delay their publication as loop closure candidates. As a result of the search, the list of the $j$ most similar images $C_t = \{I_{s_1},\ldots,I_{s_j}\}$, ordered by their associated scores $s(I_t, I_k)$, is obtained. \change{The range of these scores highly depends on the distribution of the visual words and varies between even consecutive and similar images}. Therefore, they are normalized, using a min-max technique, as follows:
\begin{equation}
  \tilde{s}(I_t, I_k) = \frac{s(I_t, I_k) - s(I_t, I_{s_1})}{s(I_t, I_{s_j}) - s(I_t, I_{s_1})}\,,
  \label{eq:minmax}
\end{equation}
where $s(I_t, I_{s_1})$ and $s(I_t, I_{s_j})$ are, respectively, the minimum and maximum scores obtained from the image search. This normalization step maps the scores to the range $[0, 1]$. Next, we discard those images whose normalized score $\tilde{s}$ is below a predefined threshold $\tau_{im}$, generating the final ordered list of image matches $\tilde{C_t} \subseteq C_t$. \change{Note that this threshold determines the number of candidates. Setting $\tau_{im}$ to a low value results quite convenient since, in this way, there are still enough candidates but the worst choices can be discarded.}

\subsection{Dynamic Islands Computation}
\label{ssec:llcislands}
In previous works~\cite{Garcia-Fidalgo2014ETFA,Garcia-Fidalgo2017}, we relied on discrete Bayes filters to detect loop closures. As it is well known, these techniques lead to increasing computational times as more images are processed, specially due to the cost of calculating the transition model. To overcome this problem, iBoW-LCD introduces the concept of \emph{dynamic island}, as an extension of the idea of \emph{island}~\cite{Galvez-Lopez2012} that locally adapts the size of the group of images. The innovation against the original concept of island is twofold: on the one hand, iBoW-LCD does not compute islands using all the previous images but makes use of a filtered set of similar images $\tilde{C_t}$ resulting from the previous step; on the other hand, the size of the islands is not fixed but depends on the similarities between neighbouring images and the camera velocity, what adapts the resulting islands to the image stream.


\begin{figure}[tb]
\centering
  \begin{tikzpicture}[scale=1.0, every node/.style={transform shape}]
    \matrix(M)[matrix of nodes]
    {
      $I_0$ & $I_1$ & |[draw,circle,fill=gray!20]| $I_2$ & $I_3$ & $I_4$ & |[draw,circle,fill=gray!20]| $I_5$ & $I_6$ & $I_7$ & $I_8$ & |[draw,circle,fill=gray!20]| $I_9$ & $I_{10}$ & $I_{11}$\\
    };

    \draw[thick,decorate,decoration={brace,mirror}]($(M-1-1.south)+(0,-0.25cm)$)--($(M-1-4.south)+(0,-0.25)$)node (A) [below,midway,yshift=-3pt,font=\small]{$\Upsilon^0_3$};
    \draw[thick,decorate,decoration={brace,mirror}]($(M-1-5.south)+(0,-0.25cm)$)--($(M-1-7.south)+(0,-0.25)$)node (B) [below,midway,yshift=-3pt,font=\small]{$\Upsilon^4_6$};
    \draw[thick,decorate,decoration={brace,mirror}]($(M-1-9.south)+(0,-0.25cm)$)--($(M-1-12.south)+(0,-0.25)$)node (C) [below,midway,yshift=-3pt,font=\small]{$\Upsilon^8_{11}$};
  \end{tikzpicture}
  \vspace{-3.0mm}
  \caption{\change{A simple example comprising 3 islands ($\Upsilon^0_3$, $\Upsilon^4_6$, $\Upsilon^8_{11}$) of different sizes, resulting from a sequence of 12 images ($I_0, \ldots, I_{11}$). Note that $I_7$ does not belong to any island. Grey circles denote the island's representative, which is the image with the highest score $\tilde{s}$ and hence the island's origin.}}
  \label{fig:islands}
\end{figure}
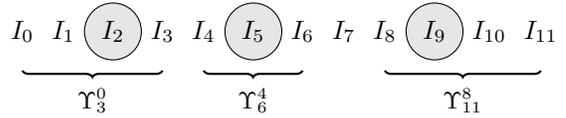
\change{In this work, an \emph{island} is defined as a group of similar images whose timestamps lie between two different instants. This criterion allows us to group images close in time and avoid them from competing to each other as loop candidates.} We denote $\Upsilon^m_n$ as the island which groups the images whose timestamps are between $m$ and $n$. Additionally, there always exists a representative image for each island, which corresponds to the image with the largest score $\tilde{s}$ within the range $[m, n]$. To manage the set of islands, images in the list $\tilde{C_t}$ are considered sequentially: for each image $I_c \in \tilde{C_t}$, we assess if its timestamp lies within the time interval of an existing island $\Upsilon^m_n$; if this is the case, the image is associated to $\Upsilon^m_n$ and the time interval of $\Upsilon^m_n$ is updated to accommodate $I_c$ and the $b$ previous and posterior frames; otherwise, an island is created, with a predefined initial size $2b+1$ around time $c$, and $I_c$ is associated to the new island. After processing all the images in $\tilde{C_t}$, the limits of the resulting islands are revised and truncated, if needed, in order to obtain a disjoint set, \change{avoiding time overlaps among islands}. \change{Figure~\ref{fig:islands} illustrates the concept through a simple example.} For each island, a global score $G$ is computed as:
\begin{equation}
  G(\Upsilon^m_n) = \frac{\displaystyle \sum_{i=m}^{n} \tilde{s}(I_t, I_i)}{m - n + 1}\,,
  \label{eq:island}
\end{equation}
which is the average of normalized scores of the images associated to the island. Finally, the resulting list of islands $\Gamma_t$ is sorted according to their global score $G$. The full process of building islands is summarized in Alg.~\ref{alg:buildislands}.
\begin{algorithm}[tb]
\caption{Building dynamic islands}
\label{alg:buildislands}
\begin{algorithmic}[1]
\Require{$\tilde{C}$: Ordered list of similar images}
\Ensure{$\Gamma_t$: Ordered list of islands at time $t$.}
\State $\Gamma_t \leftarrow []$
\For{each image $I_c$ in $\tilde{C}$}
  \State found $\leftarrow$ false
  \For{each island $\Upsilon^m_n$ in $\Gamma_t$}
    \If{$m < c < n$ and not found}
      \State associateToIsland($I_c, \Upsilon^m_n$)
      \State changeIslandSize($\Upsilon^m_n, c, b$)
      \State found $\leftarrow$ true
    \EndIf
  \EndFor
  \If{not found}
    \State $\Upsilon^{c-b}_{c+b} \leftarrow$ createNewIsland($I_c, b$)
    \State $\Gamma_t = \Gamma_t \cup \Upsilon^{c-b}_{c+b}$
  \EndIf
\EndFor
\State $\Gamma_t \leftarrow$ obtainDisjointIslands($\Gamma_t$)
\For{each island $\Upsilon^m_n$ in $\Gamma_t$}
  \State $G(\Upsilon^m_n) \leftarrow$ computeIslandScore($\Upsilon^m_n$)
\EndFor
\State sort($\Gamma_t$)
\end{algorithmic}
\end{algorithm}

\subsection{Island Selection}
\label{ssec:llcislsel}
At this step, iBoW-LCD selects the best matching island $\Upsilon^*(t) \in \Gamma_t$. To this end, it recalls the best island at the previous timestamp $t - 1$, $\Upsilon^*(t-1)$, and checks whether any of the islands $\Upsilon_n^m \in \Gamma_t$ overlaps with $\Upsilon^*(t-1)$. The overlapping islands are named \textit{priority islands} inspired by the observation that consecutive images should close loops with areas of the environment where previous images closed a loop, if any. If priority islands are found, the one with the highest global score $G$ is selected for the next step. Otherwise, iBoW-LCD chooses the first island from \change{the current set} $\Gamma_t$, i.e. the one with the largest score $G$, \change{which turns out to be the island most similar to the current image.}

\subsection{Loop Closure Decision}
\label{ssec:llcdes}
This stage of iBoW-LCD chooses first the representative of the selected island as the final loop closure candidate $I_f$. An epipolarity analysis is performed next for $I_t$ and $I_f$ to validate whether they can come from the same scene after a camera rotation and/or translation. To this end, we compute a set of putative matchings between $I_t$ and $I_f$ using the ratio test~\cite{Lowe2004} and find, using RANSAC, the inliers resulting from imposing the fundamental matrix model \change{to the set of feature matchings}. The loop hypothesis is accepted only if the number of inliers is high enough.

Note that, instead of this geometrical check, a temporal coherency technique could be applied here, like in~\cite{Galvez-Lopez2012,Khan2015}, to reduce the computational requirements of the approach. However, this last method tends to reduce the recall values. This option could be considered when using iBoW-LCD in a real SLAM system, where achieving high recall values are not essential and several correct loops are enough to ensure coherent maps.

iBoW-LCD also tracks the number of consecutive loops occurred at time $t$ in order to avoid the computation of the fundamental matrix on every image and speed up the process: the algorithm accepts a loop, without performing the epipolarity analysis, if a priority island is found and the number of consecutive loops at time $t$ is higher than a threshold $\tau_{c}$.

\section{Experimental Results}
\label{sec:exresults}
This section evaluates the proposed approach and compares it against several state-of-the-art solutions. An Intel Core i7-\change{6500U} (2.5Ghz) / 12 GB RAM computer was used in all experiments. \change{OBIndex2 made use of four cores to perform a search in four trees at the same time, while iBoW-LCD employed only a single core.}

\subsection{Methodology}
\label{ssec:methodology}
As usual in these cases, the evaluation is performed in terms of precision-recall metrics. Along with the curves, we are particularly interested in the maximum recall that can be achieved at 100\% precision, what implies no false positive detections missing a minimum number of loops. Avoiding these false positives becomes essential when the algorithm is employed in a full SLAM system, given that they can produce inconsistencies in the resulting maps. The following public datasets, taken under different visual conditions, have been considered for the evaluation: City Centre~\cite{Cummins2008a} (CC), New College~\cite{Cummins2008a} (NC), Lip6 Indoor~\cite{Angeli2008b} (L6I), Lip6 Outdoor~\cite{Angeli2008b} (L6O), KITTI 00~\cite{Geiger2012} (K00) and KITTI 06~\cite{Geiger2012} (K06). For benchmarking purposes, we use the ground truth provided by the original authors of each method except for the KITTI sequences, where the files provided by~\cite{Arroyo14IROS} are employed as a reference. \change{This last ground truth was created manually by the authors, labelling as long stops the time intervals where the vehicle was not in motion.}

\subsection{Algorithm Configuration}
\label{ssec:paramconf}
In order to find a suitable set of parameters for iBoW-LCD, we initially executed the algorithm against the City Centre dataset several times, modifying the parameters until obtaining the best possible recall at 100\% precision. The resulting parameters, which are enumerated in Table~\ref{tab:params}, have then been used for the remaining experiments. Furthermore, a collection of ORB interest points~\cite{Rublee2011} have been detected and described for each image. Note, however, that our algorithm is descriptor-agnostic and that any other binary descriptor could be used instead.
\begin{table}[tb]
\caption{Parameters values and section where they are defined.}
\label{tab:params}
\vspace{-3.0mm}
\begin{center}
\begin{tabular}{|c|c||c|c|}
  \hline
  \textbf{$K$ (Sec.~\ref{ssec:muja})} & 16 & \textbf{$P_f$ (Sec.~\ref{ssec:vvupd})} & 2\\
  \hline
  \textbf{$S$ (Sec.~\ref{ssec:muja})} & 150 & \textbf{$P_o$ (Sec.~\ref{ssec:vvupd})} & 2 \\
  \hline
  \textbf{$T_i$ (Sec.~\ref{ssec:muja})} & 4 & \textbf{$\tau_{im}$ (Sec.~\ref{ssec:llccands})} & 0.3\\
  \hline
  \textbf{$p$ (Sec.~\ref{ssec:llccands})} & 50 & \textbf{$b$ (Sec.~\ref{ssec:llcislands})} & 5\\
  \hline
  \textbf{Features per image} & 1000 & \textbf{$\tau_{c}$ (Sec.~\ref{ssec:llcdes})} & 20\\
  \hline
\end{tabular}
\end{center}
\end{table}

\subsection{General Performance}
\label{ssec:genperf}
\begin{figure}[tb]
  \centering
  \includegraphics[width=0.6\columnwidth]{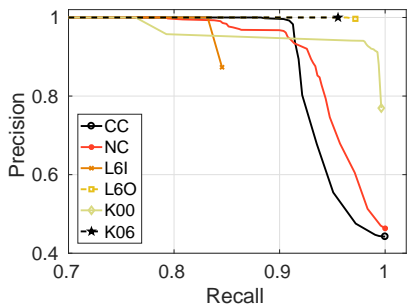}
  \vspace{-3.0mm}
  \caption{Precision-recall (P-R) curves. \change{($P = 1$ for $R < 0.75$ in all cases.)}}
  \label{fig:prcurves}
\end{figure}

\change{As a measure of general performance of iBoW-LCD, we have computed for all datasets including CC the precision-recall curves shown in Fig.~\ref{fig:prcurves}, resulting from modifying the threshold on the number of inliers required to accept a loop} (Sec.~\ref{ssec:llcdes}). As can be seen, iBoW-LCD works reasonably well in all cases, achieving high recall rates while keeping precision at 100\%. From the curves, it can be observed that the approach exhibits a very stable behaviour \change{especially for the L6O and K06 datasets}, where precision decreases minimally as recall values increase. \change{This behaviour repeats for all datasets, even under viewpoint and illumination changes. We hypothesize that deleting unstable descriptors, as done by iBoW-LCD, favours keeping more stable visual words in the dictionary, improving the general system tolerance as for the aforementioned appearance changes.}

\begin{figure}[tb]
  \centering
  \begin{tabular}{cc}
    \includegraphics[width=0.45\columnwidth]{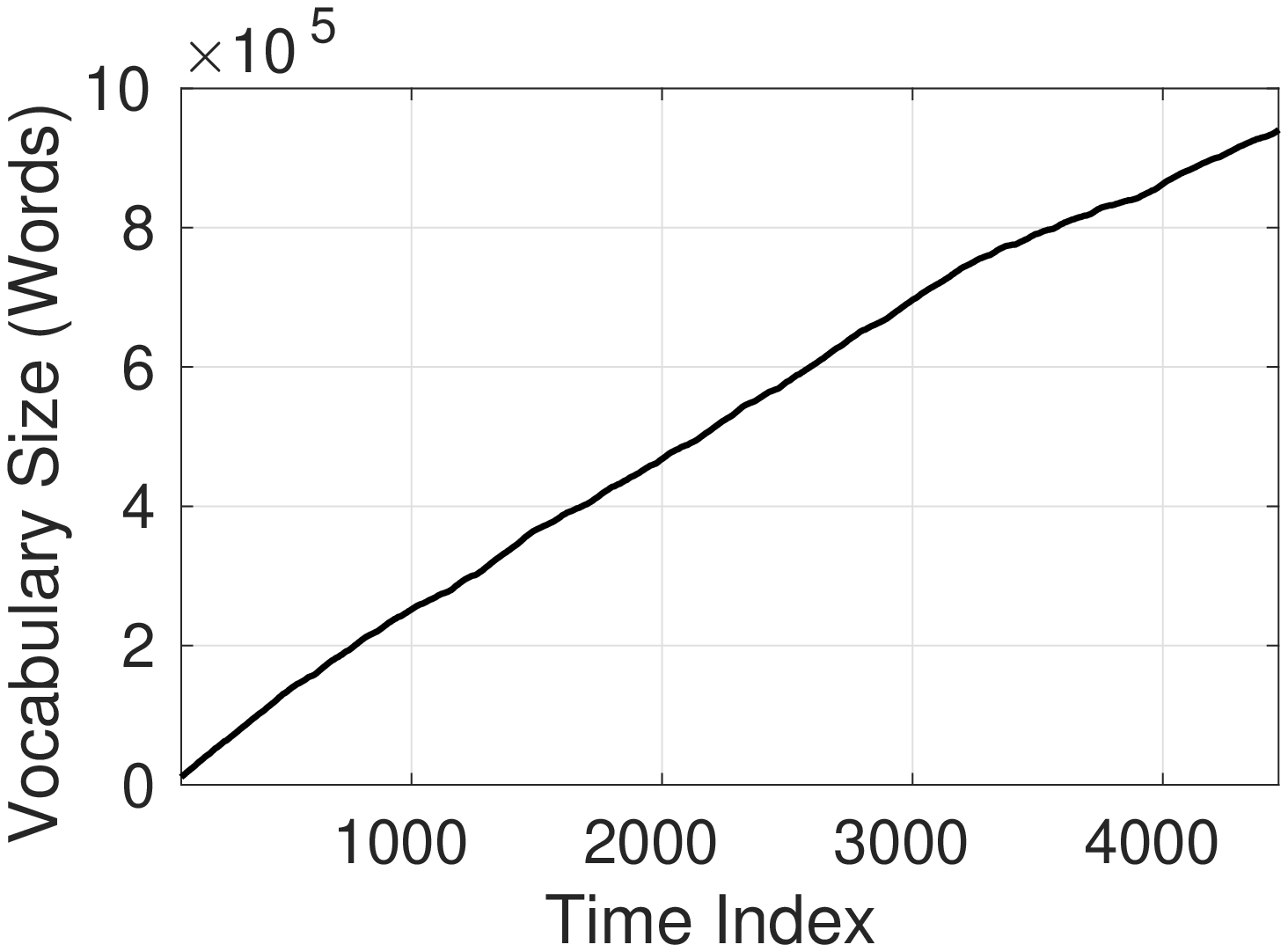} &
    \includegraphics[width=0.45\columnwidth]{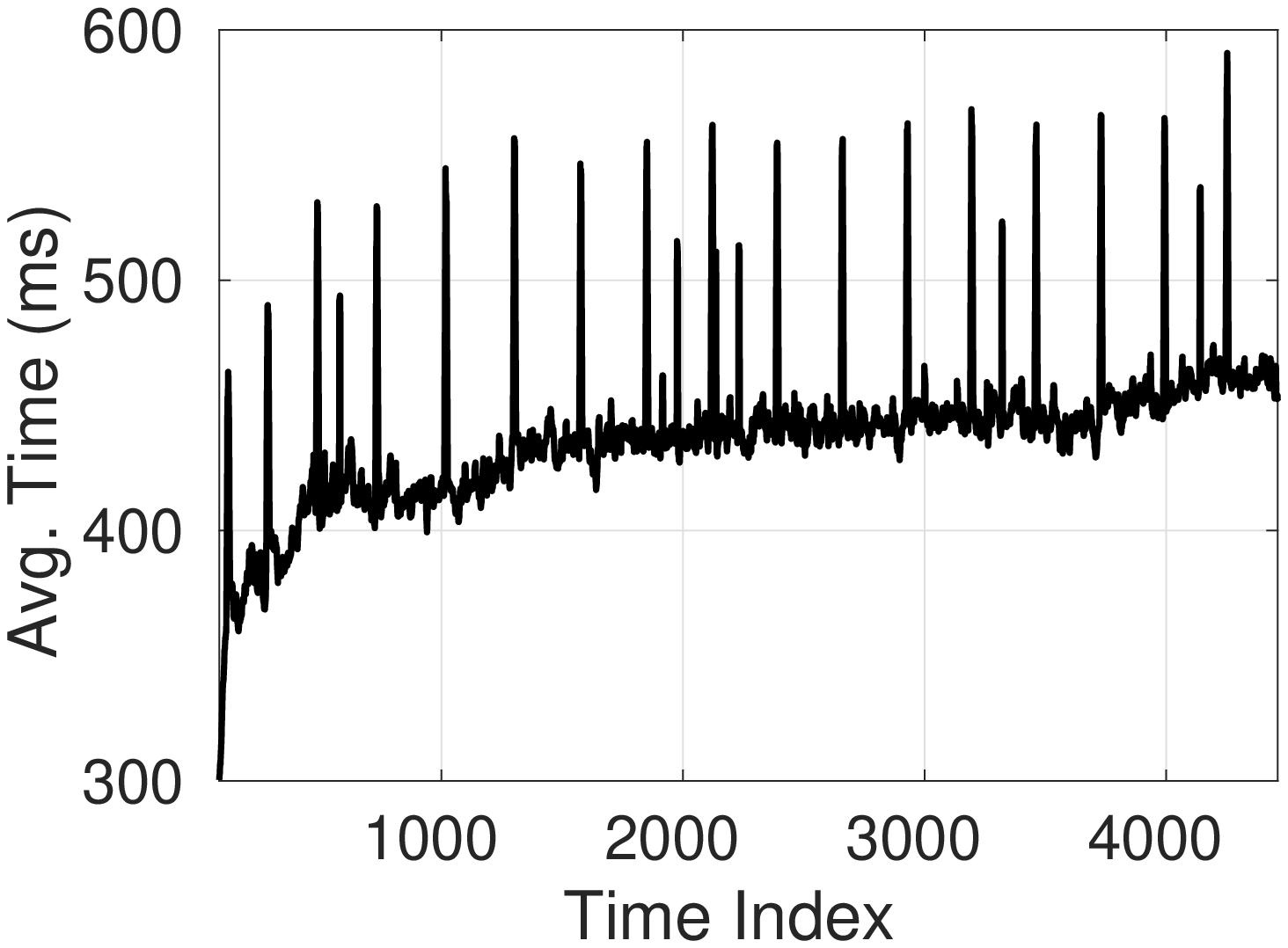} \\
  \end{tabular}
  \vspace{-3.0mm}
  \caption{Performance metrics computed over the K00 dataset. (left) Vocabulary size with regard to the number of images processed. (right) Average response time per image with regard to the number of images processed. \change{Peaks are mainly due to the rearrangement of the visual words.}}
  \label{fig:timeindex}
\end{figure}

\vspace{-0.2mm}
Next, we chose the largest dataset considered in this work (K00) to analyse the evolution of the visual dictionary size and the corresponding average response time per image, \change{which includes all the stages of the algorithm (visual word handling, image query, islands computation and loop closure decision)}. The results obtained can be found in Fig.~\ref{fig:timeindex}. \change{In spite of the fact that the number of visual words grows as more images are processed}, the average response time remains more stable, exhibiting a moderate increment. \change{Note that this growth is highly related to the trajectory performed by the vehicle: the more similar areas are revisited, the less new visual words are added, since more visual words are matched against the visual dictionary.} On average, the time required to process a frame of the K00 dataset is 432.38 ms. Times could be slightly higher than the ones presented by some off-line solutions, \change{especially} due to the delay required to manage visual words. To alleviate this, several improvements could be incorporated, such as reducing the number of features per image or applying a temporal consistency check instead of performing an epipolarity analysis. In this work, we have prioritized high recall values against computational times.

\subsection{Comparison with Other Solutions}
\label{ssec:comps}
This section compares iBoW-LCD against other state-of-the-art solutions. First of all, given that the proposed approach is an evolution of one of our previous works~\cite{Garcia-Fidalgo2014ETFA}, we want to check whether the modifications proposed here represent a real improvement in terms of response time and recall. In this regard, Table~\ref{tab:compbinmap} summarizes the final vocabulary size (VS), the maximum recall at 100\% of precision (R) and the average response time per image (T) obtained for each approach and dataset. As can be observed, the impact in terms of recall is minimum and, in general, quite similar. However, iBoW-LCD is able to process an image in less time using a more reduced set of visual words in contrast to our previous solution. We believe that this fact is mainly due to the new visual word managing process and the simplification of the loop closure scheme. Notice the high reduction of the final vocabulary size in comparison with our previous approach. \change{The variability in the obtained recall values can be attributed to the high dependence of the method on the distribution of the visual words. As shown in Table~\ref{tab:compbinmap}, deleting visual words does not always imply higher recall values, but always reduces computational times and the size of the final visual vocabulary.}
\begin{table}[tb]
\caption{Comparison with our previous approach.}
\label{tab:compbinmap}
\vspace{-3.0mm}
\begin{center}
\begin{tabular}{c|ccc|ccc}
  & \multicolumn{3}{c|}{\textbf{Previous~\cite{Garcia-Fidalgo2014ETFA}}} & \multicolumn{3}{c}{\textbf{iBoW-LCD}} \\
  & \textbf{VS} & \textbf{R (\%)} & \textbf{T (ms)} & \textbf{VS} & \textbf{R (\%)} & \textbf{T (ms)} \\
\hline
\textbf{CC} & 1.6M & 88.24 & 503.65 & 95K & 88.25 & 368.41 \\
\hline
\textbf{NC} & 1.3M & 53.15 & 489.24 & 98K & 79.40 & 352.08 \\
\hline
\textbf{L6I} & 30K & 79.09 & 24.93 & 4K & 83.18 & 19.17 \\
\hline
\textbf{L6O} & 826K & 97.51 & 304.01 & 121K & 85.24 & 249.45 \\
\hline
\textbf{K00} & 4.7M & 78.73 & 546.21 & 958K & 76.50 & 432.38 \\
\hline
\textbf{K06} & 1.1M & 84.76 & 480.49 & 212K & 95.53 & 395.16 \\
\hline
\end{tabular}
\end{center}
\end{table}

Table~\ref{tab:compothers} compares the maximum recall achieved by our approach at 100\% precision in contrast to other state-of-the-art solutions. The results reported are taken from the original papers, except the ones corresponding to Cummins~\cite{Cummins2011} and Milford~\cite{Milford2012} which were obtained by ourselves in a previous work~\cite{Garcia-Fidalgo2017}. The term \emph{n.a.} means that the corresponding information is not available from any source. From this table, one can observe that the increase in processing times is compensated in terms of accuracy, since iBoW-LCD achieves a higher recall in all datasets considered. This enhancement is specially evident in the CC dataset, where it is usually difficult to attain high recall at 100\% of precision.
\begin{table}[tb]
\caption{Comparison of maximum recall at 100\% precision.}
\label{tab:compothers}
\vspace{-7.0mm}
\begin{center}
\resizebox{\columnwidth}{!} {
\begin{tabular}{c|cccccc}
  & \textbf{CC} & \textbf{NC} & \textbf{L6I} & \textbf{L6O} & \textbf{K00} & \textbf{K06} \\
  \hline
\textbf{Cummins}~\cite{Cummins2011} & 38.50 & 51.91 & n.a. & n.a. & 49.21 & 55.34 \\
\hline
\textbf{Angeli}~\cite{Angeli2008b} & n.a. & n.a. & 36.86 & 23.59 & n.a. & n.a. \\
\hline
\textbf{Milford}~\cite{Milford2012} & 68.98 & 49.39 & n.a. & n.a. & 67.04 & 64.68 \\
\hline
\textbf{Khan}~\cite{Khan2015} & 38.92 & n.a. & n.a. & n.a. & n.a. & n.a. \\
\hline
\textbf{G\'alvez-L\'opez}~\cite{Galvez-Lopez2012} & 30.61 & 55.92 & n.a. & n.a. & n.a. & n.a. \\
\hline
\textbf{Mur-Artal}~\cite{Mur-Artal2014} & 43.03 & 70.29 & n.a. & n.a. & n.a. & n.a. \\
\hline
\textbf{Bampis}~\cite{Bampis2017} & 52.36 & 74.60 & 42.32 & 49.55 & n.a. & n.a. \\
\hline
\textbf{Zhang}~\cite{Zhang2016} & 41.18 & 59.20 & n.a. & n.a. & n.a. & n.a. \\
\hline
\change{\textbf{Stumm}}~\cite{Stumm2016} & 38.00 & 39.00 & n.a. & n.a. & n.a. & n.a. \\
\hline
\change{\textbf{Cieslewski}}~\cite{CieslewskiS17b} & n.a. & n.a. & n.a. & n.a. & $\approx$60.00 & n.a. \\
\hline
\textbf{iBow-LCD} & \textbf{88.25} & \textbf{79.40} & \textbf{83.18} & \textbf{85.24} & \textbf{76.50} & \textbf{95.53} \\
\hline
\end{tabular}
}
\end{center}
\end{table}

\section{Conclusions and Future Work}
\label{sec:conclusions}
In this work, we have introduced iBoW-LCD, an appearance-based loop closure detection algorithm, which mainly relies on an incremental Bag of Binary Words scheme to retrieve previous similar images. This incremental visual dictionary builds on a hierarchical structure to efficiently search, insert and delete new visual words on line, avoiding the main drawbacks that off-line approaches present. Next, iBoW-LCD makes use of a novel concept to group similar images close in time called \emph{dynamic island}, which naturally exploits the nature of image sequences to detect loop closures. The proposed method has been validated using several public datasets, obtaining competitive results in comparison with other state-of-the-art solutions.

Referring to future work, we will consider to extend the methods developed in this paper to a hierarchical loop closure scheme, given the good results obtained in this matter in one of our previous publications~\cite{Garcia-Fidalgo2017}. \change{We will investigate other appearance-based methods to group images. To further favour the long-term operation of the method, a mechanism to mitigate the growth of the dictionary (e.g. based on response time) could be useful.} Additionally, we also plan to enhance the response time of iBoW-LCD parallelizing some of their stages. Finally, we want to incorporate our solution into a complete SLAM / 3D reconstruction framework.

\addtolength{\textheight}{-12cm}   



\bibliographystyle{IEEEtran}
\bibliography{ms}

\end{document}